# Monitorable-by-design architectures for a post-Astra world: a compartmentalization-based approach

**Jake R Watts**, **Joel Sokol**
Georgia Institute of Technology
jwatts60@gatech.edu

**Abstract:** Frontier models like GPT-6 Astra are increasingly able to do substantial reasoning without a monitorable chain of thought, and chains of thought are becoming less legible and are often monitored asynchronously to reduce cost. With the future of chain of thought monitoring unclear, it may be necessary to design LLMs with monitorability in mind. We propose a modification to existing LLM architectures based on GRAM, SGTM, DEMix, BTS, and similar methods to compartmentalize dangerous capabilities so that their use, especially in combination with dangerous reasoning types that may indicate strategic adversarial use, can be monitored more cheaply than chain of thought, and in a way that might outlast monitorable chain of thought. We propose several variants of this architecture designed to balance tradeoffs and discuss other benefits, such as capability removal.

# 1. The problem: How do we monitor as chain of thought becomes unreliable?

Chain of thought monitoring is losing reliability at the frontier, as OpenAI acknowledged after Astra. [3]

Chain of thought + action monitoring is currently the primary method labs use to monitor what their models are doing[1] [2]. Astra, for various reasons, seems to be dramatically more capable than previous models at doing complex tasks without verbalized reasoning[3] [4]. Other experiments show Astra can control its CoT far better than prior models[3] and (at least when instructed) can evade monitors on tasks where it no longer needs extensive reasoning[3]. Models often reason about being monitored[3], and when told they are being monitored, are more likely to disguise their CoT[3]. Many at frontier labs believe CoT monitoring has few available substitutes[5], though they are actively looking for alternatives[3]. At the same time, many believe neuralese/human-illegible reasoning tokens will someday dominate, on the premise that human-legible reasoning is unlikely to be the most efficient way to condense reasoning[6] [7] [8] [9].

We believe LLMs may need to be designed so that, at minimum, we can approximately discern what kinds and combinations of concerning capabilities are being utilized at each token/in a task, without relying on models choosing to reveal their thinking. We believe capability compartmentalization can achieve this, and expect the cost and capability tradeoffs to be tolerable.

This paper makes four main contributions: (1) a modular architecture intended to compartmentalize dangerous capabilities into experts and modules (Section 4.1); (2) a monitoring signal based on routing, including the co-activation of strategic and dangerous-domain components (Section 4.2); (3) training procedures intended to enforce compartmentalization (Sections 4.3–4.4); and (4) a set of proposed experiments to test the approach (Section 6). Much of this is synthesis combining existing methods and some original contributions, those original contributions (to the best of our knowledge) including: (i) the hypothesis that the most important signal of dangerous behavior is dangerous capability (cyber, bio, weaponizable chemistry and physics) co-occurring with dangerous planning that's harder to fully isolate (long-term strategy, theory of mind, game theory, games of asymmetric information, modeling deception, detection evasion, dangerous persona modeling); (ii) attempting to isolate said dangerous planning, instead of just dangerous capabilities like cyber; (iii) attempting to isolate dangerous personas as a monitoring signal or alignment technique; (iv) an architecture combining expert- and module-level compartmentalization, and the idea of commander then modules then interpreter; (v) modifications to training like exempting early and late (often syntax-encoding) layers from forced routing, enforcing compartmentalization throughout RL with task labels; and (vi) training monitors using routing signals, with many modifications of existing ideas.

# 2. Existing monitoring methods and their tradeoffs

**Table 1:** Major existing approaches to monitoring frontier models and the possible tradeoffs of each going forward

| **Approach** | **What it monitors** | **Main strength** | **Main weakness(es)** | **Status vs. Astra-type gap** |
|---|---|---|---|---|
| CoT monitoring | Reasoning tokens | Cheap, ideally legible to humans | Not necessarily faithful, models can censor it, models can choose not to rely on it, possible selection pressures against remaining human-readable | Degrading[3] |
| Training for legible CoT | Reasoning tokens | May keep CoT usefulness from degrading | Optimization pressure on CoT itself can teach obfuscation; legibility does not imply faithfulness; might tradeoff with capabilities, limiting actual adoption | Possibly[10] hazardous[11], unlikely to address latent capability[3] [12] |

| Approach | What it monitors | Main strength | Main weakness(es) | Status vs. Astra-type gap |
|---|---|---|---|---|
| Activation probes | Internal representations | Works without CoT | Probes can be evaded; need labeled examples of misbehavior; can only find what it was trained on; will likely struggle to detect risks that weren't anticipated in advance | Powerful[13] for well-anticipated[14] bad behaviors, may not generalize out of distribution[13] |
| Mechanistic interpretability | Internal circuits | Deepest insight | Struggling to scale to frontier models, difficult to verify conclusions of many methods | "Probably too popular relative to other alignment research"[15], still struggles as a method of detecting bad behavior[16] |
| Output and action monitoring | Actions, tool calls | Catches harmful actions directly | Misses intent and sandbagging until it acts, many hazards may not come from or be evident in tool calls | In production use, easy but limited to actions & tool calls |
| Capability localization for monitoring (DEMix, BTX, GRAM, SGTM)[26] [38] [31] [23] | Router signals | Structural signal, independent of CoT, might aid with capability mitigation/ unlearning | Possible leakage, capability tradeoff, limited testing at frontier scale, may need to be built-in from the start | Underdeveloped but high potential, this paper's proposal |

Capability localization may complement activation probes and mechanistic interpretability, making architecture more interpretable by default. It may also make monitoring more useful, as disabling modules when activity is concerning may be a useful alternative to classifier-triggered fallbacks to weaker models (like was implemented with Fable[17]) or a whole-model kill-switch. Importantly, the broad strokes information of which capabilities are being used may be very difficult for models trained with methods forcing compartmentalization to censor, and might be easy to interpret even in moderately adversarial settings.

Prior work has used expert routing to detect or steer harmful behavior in standard MoE models, where specialization is emergent[28] [29], and has localized capabilities into removable modules for unlearning[31] [23] [36]. [36] briefly suggests that localized components could be monitored; SafeMoE isolates unsafe knowledge into designated experts but uses routing to gate outputs rather than to monitor[39]; and GRAFT applies gradient routing during RL post-training, routing by monitor-flagged trajectories rather than task labels[40]. This paper synthesizes an architecture integrating these methods along with original ideas into something designed to maintain capabilities while maximizing the monitorability benefit of compartmentalization.

# 3. Why compartmentalization might be necessary

If latent reasoning keeps growing, monitoring will need signals the model doesn't produce voluntarily, and architecture is one place to get them.

### 3.1 The trend

For years, no-CoT capability had been increasing at about half the rate of with-CoT capability[12]. Astra seems to have bucked this trend with a massive increase in no-CoT capability[3], and CoT is getting less interpretable[10]. We expect that CoT token reasoning will get more polysemantic as more meaning is loaded onto the same vocabulary than natural English typically does, and increasing serial depth and model parameter size may continue to improve what can be done without requiring reasoning. Distillation on outputs that require reasoning might also allow especially very large models[18] to emulate the effects of reasoning without the large inference-time cost[19]. Chain of Thought monitoring must not remain our best insight into a model's thinking long-term.

### 3.2 What an alternative signal needs

Most importantly, we need something that is innately difficult for a model to voluntarily control. The signal we get about a model's use of a concerning capability must be difficult to mask without sacrificing much of that ability. It should ideally be cheap to read (ideally does not require training domain-nongeneralizing interpreters or probes[13] or the use of an LLM reading CoT transcriptions (with monitoring overhead in some cases 20% of inference cost[20])) so that labs will actually use monitors all the time (which might have prevented the Hugging Face incident[21] and/or related incidents). At minimum, it should signal when the model is doing something that might be concerning and provide some insight into a model's intent.

### 3.3 Why architecture fits

We find little evidence that architecture-based monitorability has seen nearly the same level of investment as CoT monitoring or standard mechanistic interpretability, and therefore the marginal return on investment from moderate investment from labs may be substantial. CoT monitoring is gameable because models are usually trained in a way that does not ensure their dependence on interpretable CoT[22]. Probes struggle to generalize out of distribution, and other post-hoc methods might be made less challenging and more effective than they currently are if we designed LLMs to be interpretable (or at least monitorable) from the ground up. Architectures designed to be monitorable may reduce the cost of monitoring by not requiring LLM monitors (also reducing reliance on LLM monitors that could lie). Compartmentalization may also have additional benefits to unlearning[23].

### 3.4 Why now

The success of MoE[24] [25] shows conditional computation is cheap and doesn't sacrifice capability. That many compartmentalization methods have been developed solely because they may benefit capabilities and/or inference-time cost lends additional strength to this argument. Labs currently fear that as capabilities are starting to become genuinely concerning they're losing the ability to effectively monitor the thoughts of the next gen of LLMs[3]. Now more than ever, labs might be willing to design LLMs in a way that makes this inherently less of a problem.

# 4. The proposal: Commander and Modules for Monitorable Compartmentalization (CoMMCo)

A generalist commander with compartmentalized MoE which can then route to one/k of several (dangerous) specialist modules or a generalist module, compartmentalizing dangerous capabilities into specific experts and modules for monitorability and capability removal.

This is one proposal for an architecture we believe would provide a high degree of monitorability, while attempting to maximize capability and minimize cost. We expect labs will modify this heavily before implementing it, using proprietary methods we cannot anticipate.

### 4.1 Architecture

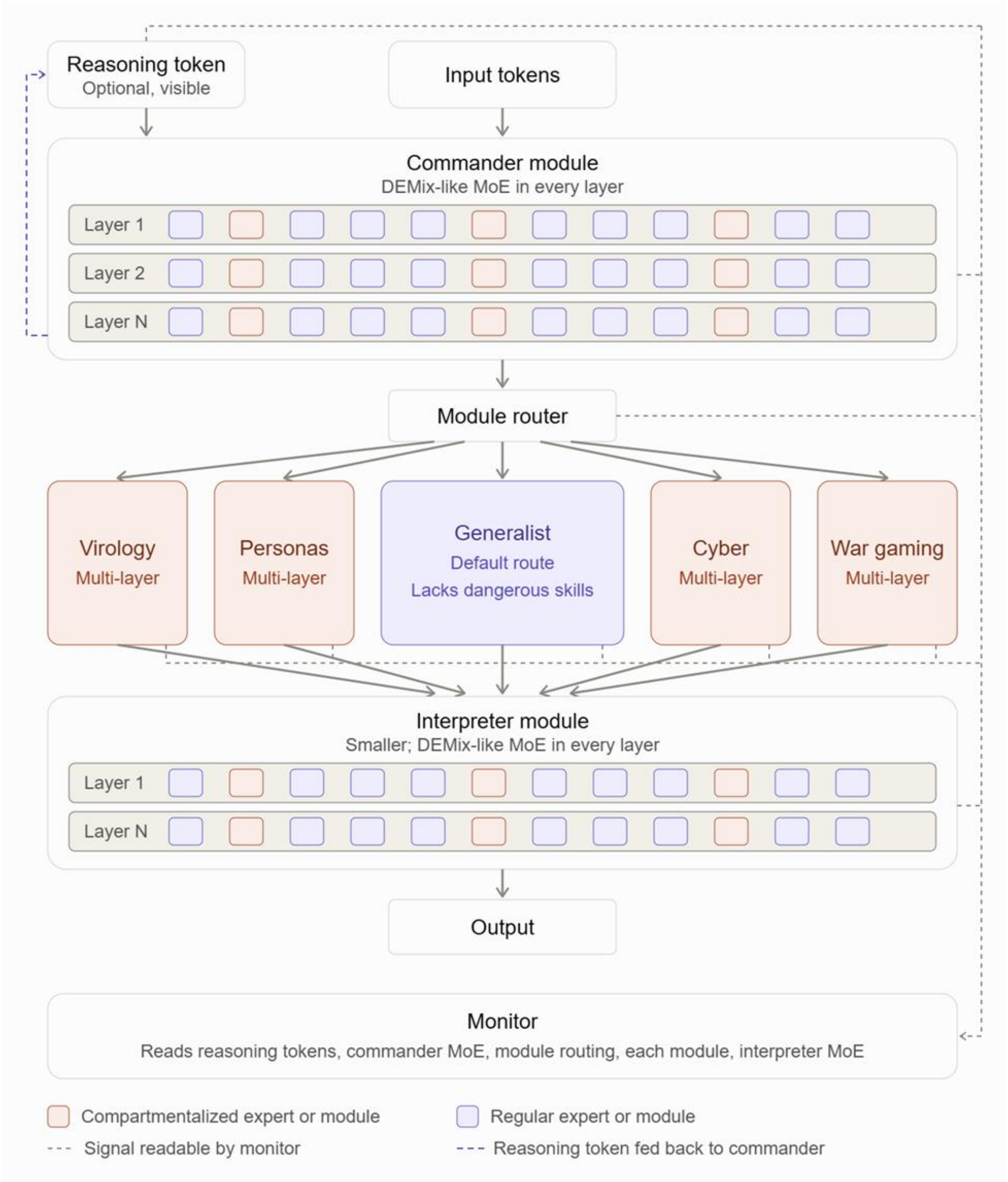

**Figure 1:** The proposed architecture. Inputs pass through a commander module whose MoE layers mix compartmentalized (labeled) and regular experts. A module router sends each token to one or more generalist modules or to compartmentalized specialist modules, whose outputs are then processed by some shared layers similar to the commander. A monitor reads routing at every stage to detect potentially dangerous patterns.

In the following, “expert” refers to FFN-level experts in the MoE sense, while “module” refers to something with multiple transformer layers.

We imagine that the model contains some commander module similar to ordinary MoE transformers (Fig. 1). While it could be more monitorable to have no such commander, we believe this will give a better balance with capabilities, as this commander would be where planning and routing to future specialists emerges and might be useful in maintaining generalist capabilities.

To ensure that some meaningful signal still exists that could alert a monitor to dangerous capabilities, we suggest a DEMix[26]-like MoE (modified with top-k routing) at each transformer block to compartmentalize at least dangerous capabilities (especially those that might be implicated in many generalist tasks, like long-term planning, setting sub-goals, game theory, theory of mind, deception, adversarial settings, detection evasion, etc.) where experts are exposed to different subsets of training data and encouraged to specialize in these capabilities. Non-dangerous capabilities may not need to be compartmentalized, so many experts with little interpretable meaning might also exist. Some top-k router should live in front of each of these sets of experts as might already be commonplace in frontier LLMs.

(It could be that simply adding this much to existing architectures is enough to dramatically increase monitorability for dangerous capabilities. If so, we'd suggest things like virology and cyber should be a focus for compartmentalization in these experts.)

Then, following the commander module, some number of more specialized modules should live downstream. A router should route signals through one or more of these specialized modules. Perhaps these modules should be able to exchange information between each other (either by allowing different specialists in each layer like in BTX[38], or via BTS-like stitching[27]) or perhaps not (we will discuss what we believe may be pros and cons of each later). Dangerous specialized capabilities should be localized into these modules, like virology, cyber capabilities, modeling of evil personalities, war gaming, deception, etc. These should extend for multiple layers, unlike the experts in the commander layers. At least one of these modules should be a "generalist" lacking only in dangerous capabilities, through which the majority of tasks are routed (not unlike the hub in BTS[27]). Perhaps instead/additionally the commander should have the option of routing to no specialized modules, and receive a tiny reward in training for saving inference cost when doing so.

A second commander-like "interpreter" module should follow these parallel modules before the final output as well (not unlike in BTS[27]), though it may not need to be as large as the pre-specialist commander. It may also make sense to give it DEMix-like MoE like the pre-specialist commander so that it too can be monitored for any dangerous capability elicitation.

The commander at each token might instead elect to output a visible (ideally interpretable) reasoning token that is fed back into itself without eliciting the downstream experts.

### 4.2 The monitoring signal

A monitor gets to see which experts and which modules are being used at each token, and can flag or even halt if things like game-theoretic scheming and virology, for instance, are both being elicited at once.

Once this architecture is set up and trained, if compartmentalization is at least partially successful, the following should provide insight into what the model is doing. If the model is making heavy use of, say, a virology module (or, if compartmentalization is good enough, perhaps if it is making any use of a virology module), that should signal to a monitor that the LLM is probably thinking about virology. Perhaps this should always flag that additional scrutiny should be paid to its activities, or that the model should have to request permission for tool calls it would not otherwise (or even to continue thinking at all). We will not speculate too much as to what such policies a lab might implement, only that this architecture enables many that could be difficult otherwise.

The same may be true of the smaller FFN-level experts either within modules (which might have further-compartmentalized FFN-level experts, or not) or the generalist or interpreter. This idea has been demonstrated successfully before even when expert specialization was merely emergent[28]. The signal might be a bit harder to read, as each token would be routed through multiple experts at different layers, which might have different labels (genuine deception may only require routing through just one or two deception experts, even if hundreds are available, and it may be that activities not obviously related to deception are routed through many deception experts innocently for emergent reasons, or for interpretable reasons like assessing if something it's reading online is true). However, even this somewhat muddy signal, with a bit of post-training experimentation, could provide useful insight if one assessed what the activations tended to look like when the model was doing/considering something actually concerning, which might be more useful than existing monitoring methods and/or enhance existing methods by making the underlying activations more interpretable.

In particular, the combination of strategic reasoning with dangerous capabilities, such as routing to deception, evasion, theory of mind, and long-term planning experts in the commander, with cyber capabilities or virology modules throughout a task at higher-than-normal levels could be useful in triggering additional safety measures. If, as we suspect is the case, the highest risk comes from the use of dangerous capabilities while trying to evade detection and as a part of a long-term plan, it may be hard for models to hide this activity without being ineffective at it if compartmentalization is successful, and these particular combinations of monitoring signals may be very rare in benign cases, especially if the benign task does not obviously require those specialist modules.

When such signals are received, the dangerous modules might be disabled (by not routing through them even when they are in the top k), and possibly even certain FFN-level experts should be disabled. This might even be done to drive behavior towards safety, which can be effective even when expert specialization is emergent-only[29]. Alternatively, these signals could disable certain tool calls, send warnings to users or staff, impose reasoning limits, require additional permissions, route to a weaker model, initiate expensive safeguards like in our prior work[30], or even trigger a kill switch. Or they may be used only in post-hoc investigations to discern where things went wrong and how bad behavior could have been detected.

These signals might also improve CoT monitoring, by training a monitor on what combinations of tokens tend to elicit dangerous capabilities the most, or be produced the most when dangerous experts/ specialists are used. They might similarly help train probes, SAEs, and other interpretability tools.

### 4.3 Training procedure

For the best results, the different FFN experts and specialist modules should be trained in such a way that compartmentalization is actually achieved. Best practices will almost certainly need to be determined experimentally, but here we describe below possibilities based on/ expanding on existing methods.

In constructing base models, data might be labeled with one or more capabilities that data relates to that correspond to labeled modules or experts. During base model training, routers might be required to route through capabilities matching the label. This might be only initially, with emergent routing preferences arising after an initial window, or through all of training, with router training happening in a later step in which experts/specialists are frozen, or with routing forced stochastically throughout training. Off-label experts/modules might have their weights frozen, or updated less, or have a KL penalty or weight drift cap. Routers might receive an update driving them to have routed to the corresponding experts/modules. Additionally, GRAM-like gradient routing[36] [31], selective gradient masking[23] or including a loss term to penalize high-performance from off-label modules/experts to help force compartmentalization, should prevent leakage by limiting learning and capability of off-label experts/modules.

Additionally, many specialist modules might be trained only after the commander, interpreter, and generalist modules have been trained, with masking and other methods preventing them from acquiring specialized knowledge.

In RLHF, RLVR, and other task-based fine-tuning, tasks and even sub-tasks could be labeled (either before the task is given to the model, or based on how the model broke up the task) based on what capabilities they should require. Here, the same set of options might be useful:

- Selectively freezing, stochastically freezing, or increasing KL loss/decreasing divergence maximum, on off-label modules/experts
- Fixing routing during certain stages of training and/or a loss term on routers based on correspondence to the labeled task.
- GRAM- or SGTM-like gradient routing or masking[31] [23] [36] or penalties for high performance from off-label modules/experts
- For some specialized training, like cyber-related RLVR, perhaps everything but the related specialist module(s) should be frozen for all or most of training, or its generalist alternatives be post-trained (a bit) with gradient ascent, to ensure non-cyber modules lack cyber capabilities.

Such measures may be most effective if different tasks are mixed in throughout training, rather than very long batches of highly specialized training, though KL terms in loss (or limits on total deviation) may be enough.

Early specialization might actually be sufficient, as a positive feedback loop between routing and specialists following early specialization might be enough to induce dependence on specialized modules and prevent substantial leakage[36] [23]. This should be very carefully studied before relied on. We would recommend that some degree of enforcement of compartmentalization be utilized to various degrees throughout the entirety of training, followed by evaluations measuring compartmentalization, such as demonstrating that routing activations can be used to determine tasks, and that removing modules/experts significantly reduces dangerous capabilities without destroying other capabilities.

### 4.4 Additional training considerations

Care should be taken that models are never penalized in training based on readings from the monitor in a way that might train the model to censor the signals the monitor relies on[11]. Routing-based penalties should be based on correspondence to data/task labels, not monitors detecting unsafe behavior (even/especially if this is the best means of detecting unsafe behavior).

Load balancing of MoE routing should exempt dangerous capabilities. If multiple experts are labeled with the same dangerous capability (or otherwise have overlapping labels, if labeling is more complex), then load balancing should be only between experts that share labels.

Importantly, based on $L^3$ [32], it may be important that safety training and refusal not be compartmentalized, especially in open-weight models. Safety training might use different training techniques than those outlined above (possibly including different load balancing policies, or different routing during training) to ensure refusal/safety is not concentrated. During such training, care should be taken to ensure there is no capability signal in the reward that might cause dangerous capabilities to leak out of their modules/experts. There may be particular risk if cheating is not caught in safety testing, as scheming-related capabilities may be rewarded and leak into other experts.

For the benefit of capabilities, it may be that experts in the commander and interpreter should not have forced routing during training (at least in very early and late layers) so that syntax specialization[33] can develop without making the smaller numbers of dangerous modules syntactically incoherent. Leakage should then be prevented (at least in these layers) through the other means listed, e.g. freezing.

### 4.5 Alternate uses

The proposed architecture might also be useful in unlearning by removing dangerous modules/experts entirely. This is similar to the selective gradient masking idea, which has been demonstrated to work[23] [31].

Additionally, this may also be useful in alignment. During various stages of training, but perhaps especially in construction of the base model, dangerous personalities might themselves be compartmentalized if labeling their appearance in training data is feasible. This might make steering and alignment easier by removing them or training routers to avoid them most of the time, or might provide a signal during monitoring of possibly concerning behavior. Though we discuss this only in this subsection, personality compartmentalization might ultimately prove to be the best use case of this architecture if effective. This might be amplified using a loss term penalizing off-label modules for accurately predicting tokens from such personalities. This builds on existing work related to emergent personas within LLMs[34] [35].

Safety note: For open weight models, there is a risk that compartmentalization might make jailbreaking via $L^3$-like methods[32] easier. For dangerous persona compartmentalization, routing to a dangerous persona might be a one-step jailbreak. Perhaps, then, weights for such modules should not be disclosed publicly. For closed-weight models, this could still be an issue if untrusted users/agents are given access, especially if they can fine-tune routing.

### 4.6 Tradeoffs

- Monitorability vs. capability tax: The existence of commander and interpreter modules poses some risk of dangerous capabilities residing in these layers (though various training methods may mediate this greatly, such as freezing these layers during dangerous capabilities training, or applying a loss term in these layers as well for high capability when dangerous modules are unused or absent). However, their presence may offset capability tax greatly by ensuring capability compartmentalization at the level of modules doesn’t get in the way of generalist capabilities, like synthesis across domains or planning involving a mix of abilities. Expert specialization within these layers could keep even these layers somewhat monitorable. There is some risk that reducing the amount of data/training that non-dangerous generalist modules get by freezing them (or similar) during dangerous capability training lowers their capabilities. This might only be a substantial problem if a large fraction of data/ training is of dangerous capabilities.

- Monitorability vs cost: The fact that, unless non-sparse stitching between modules is used, only some modules are used for each token should lower inference-time cost the same way that MoE does. However, it might need to increase total parameter count to match performance of other approaches, and data labeling might impose some training-time cost as well. Similar approaches seem to have capability downsides equivalent to only a few percentage points in cost[23], so a lower inference-time cost alone might be enough to justify this.

### 4.7 Variants

A key question concerns the setup of the modules. Should a router choose the top 1 or top k modules the same way routers choose experts? Choosing just one limits the ability of dangerous capabilities to talk to each other and produces an extremely interpretable signal of what was most-considered in producing that token, but might make such specialist modules weaker at things like planning, which while safer, may be a capability cost companies are unwilling to pay for things like cyber capabilities.

Alternatively, BTX-like mixing or BTS-like stitching[38] [27] could allow these modules to better "talk to" the generalist module(s), which would probably improve multi-faceted capabilities. However, this could muddy the signal for monitoring, especially if stitching is not sparse. One might overcome this by adding a bias term (routing prior—not training penalty) against routing to a dangerous module if the above layer made no use of it (to ensure dangerous modules don't participate in virtually every token all the time like they otherwise might in BTS[27]).

Instead of a single generalist, one may have multiple harmless modules stitched together like in BTS[27], with dangerous capabilities compartmentalized and routed to or stitched in a sparser way to preserve monitor signal.

We note that stitching between an unsafe module and another module may introduce complexities in GRAM, SGTM[31] [23], gradient ascent, etc., which might either make doing so undesirable or require modifying these methods.

# 5. Limitations and open questions

- **Leakage:** A capability may live in shared parameters, so the model can use it without activating the flagged module. However, GRAM and SGTM both have demonstrated success in compartmentalizing knowledge (at least in small models)[31] [23]. The use of modified loss functions that punish dangerous performance without querying the relevant experts should also help minimize leakage into shared parameters, as should freezing and other methods discussed in this paper. Further, even when dangerous capabilities are not fully contained, models might rely (or attempt to rely) on specialized modules/experts to maximize expected reward (or even simply because they were trained to)[36].
- **Composition:** Dangerous capabilities may be built from benign ones (e.g. general chemistry plus planning), so no monitor can flag their use. This is a real concern, especially if companies were to rely on this approach for unlearning without other mitigations. For monitoring, however, we expect it would be much more difficult for a model trained as described to—without further training—become highly competent at combining benign capabilities related to, say, chemical weapons while also never thinking about them in a way that causes routers to use the appropriate module, than it would be to just use the module. Further, while an AI that had never seen anything about, say, virology in training could still teach itself virology from the internet, it could be that tool call monitoring would likely be effective at detecting this, and LLMs can struggle at on-the-job learning.

- **Erosion under RL:** The same RL dynamics that eroded CoT monitorability[3] could push capabilities out of their compartments. As noted in section 4.3, freezing, masking, gradient routing[23] [31] [36], KL loss, and periodically doing a bit of gradient ascent on off-label capabilities during RL should make this much less likely. It is true that RL could be conducted in a way that undermines compartmentalization, so labs will need to develop best practices to avoid this. If this tends to be a problem, evaluations could be conducted throughout RL to see if leakage is happening and trigger reversion or unlearning (though, to prevent sandbagging[3] from undermining this, we would suggest focusing on preventing this rather than trying to detect and fix it).
- **Shallow specialization:** Experts in standard MoE models often specialize on surface token features, not semantic skills[33].Most non-specialized experts probably still will, which is why we suggest forcing compartmentalization in multiple ways (Monet shows experts can be made near-monosemantic by design[37]). We also address this somewhat in section 4.4, where some layers are allowed to develop syntactic understanding.
- **Capability tax:** Frontier AI companies will be reluctant to adopt a design that costs meaningful capability. We have attempted to design an architecture to minimize anticipated capability tax, including some variants, like the BTS variant, that we expect to trade some interpretability gains for capability. Ultimately labs will have to assess if and how variants of this idea can preserve capabilities, if in fact they are unwilling to trade capability for safety. Methods like SGTM have had taxes on the order of 6%[23], so we suspect similar might be possible here, and the success of BTS and MoE[27] [24] [25], and the possibility of sparse routing reducing inference time costs gives us confidence that per-inference-cost capabilities should not be unbearably diminished after the best variants of this architecture have been discovered.
- **Adoption:** This method requires changes to pretraining or architecture, not a bolt-on. This is a significant downside. We believe there may be ways of bolting this on to existing models (e.g. BTX- or BTS-style branching from a pretrained model[38] [27]), especially at the MoE level, but it will likely require a bit of unlearning on the off label modules, and will likely be less effective at compartmentalizing than building it from the ground up. We still believe the urgency of the issue merits building this from the ground up, but perhaps people should start testing bolt-on versions of this method now so that, if it works, adoption is an easier ask.

# 6. Proposed experiments

**Table 2:** Experiments which might provide insight into the promise of the proposed architectures and methods.

| # | Question | Minimal setup | Key metrics | Supports the idea if… |
|---|---|---|---|---|
| 1 | Does module activation detect capability use with no CoT? | Small modular transformer trained from scratch on synthetic tasks | Monitor recall vs. Monitor with only output baseline w/ CoT suppressed, observe false positive rate | Monitor's recall clearly above baseline, false positives not intolerable |
| 2 | Is the capability localized to the module? | Compare model to model with specialist ablated | Accuracy on some related benchmark before vs. after ablation | Specialist skill degrades, off-specialist abilities survive |
| 3 | What's the capability tax? | Compare to standard MoE baseline with similar inference or training cost | Task performance gap on multiple benchmarks | Small gap (ideally <10%, but a bit higher may be fine at first, as the method will be improved upon over time). |
| 4 | Does localization survive RL to avoid the module? | Fine-tune with a penalty on routing to the labeled expert | Whether the skill disappears or relocates | Skill degrades rather than relocating |
| 5 | Can a pre-trained model be cheaply converted? | BTX-style branch of a small open model[38], train one expert, merge | Repeat experiments 1–4 | If the results hold here too, this may be much easier to adopt. |

# 8. Conclusion

Amidst increasing concern that chain of thought monitoring may not be a long-term solution for monitorability, we propose that LLM architectures be designed for monitorability. We propose that compartmentalization of dangerous capabilities and risky strategic reasoning might produce an important and hard-to-hide monitoring signal with spillover benefits in unlearning. We give an example of an architecture and training methods designed to maximize monitorable compartmentalization at low cost to capabilities, and provide several variants with different expected tradeoffs. We propose a number of experiments to test this, particularly if the architecture actually produces compartmentalization, if that actually allows monitors to detect dangerous behavior when used, and to estimate how severe the tradeoff with capabilities may be.

# Acknowledgments

AI tools were used for literature search, formatting, generation of figure 1, citation checking, and editing feedback.